\documentclass{article}
\usepackage{spconf,amsmath,graphicx}
\usepackage{booktabs}
\usepackage{hyperref}
\usepackage{multirow}

\title{U$^2$Mamba: A Two-level Nested U-structure Mamba for Salient Object Detection}
\name{\begin{tabular}{c}
Junhui Li$^{1,2}$ \quad
Jialu Li$^{3}$ \quad
Youshan Zhang$^{2*}$
\end{tabular}\thanks{This research was funded by the Research
Project of Chuzhou University (Grant No. 2025qd36).}}

\address{
$^{1}$School of Electronics and Information Engineering, University of Science and Technology Liaoning,  China\\
$^{2}$School of Artificial Intelligence, Chuzhou University, Anhui, China\\
$^{3}$MS in Artificial Intelligence, Yeshiva University, New York, USA\\
youshan\_zhang@chzu.edu.cn, *Corresponding Author
}
\begin{document}
%
\maketitle
\begin{abstract}
Mamba-based models have emerged as a promising alternative for salient object detection (SOD), offering significant advantages in modeling long sequences.  However, existing models often fail to explore contextual information and the depth of the entire architecture. This paper introduces U$^2$Mamba, a powerful and innovative U-structured network for salient object detection. We propose multiscale Mamba U-blocks (MMUBs) that enhance the model depth to improve local feature extraction capabilities. Our newly developed nested U-structure, incorporating MMUBs, enables the network to integrate various receptive fields from shallow and deep layers, thereby collecting richer contextual information and longer-range data without being constrained by resolution. Instead of using the traditional deep supervision scheme and top-level supervised training, we propose a hierarchical training supervision method where the loss is computed at each level during the training process. Extensive experiments demonstrate that U$^2$Mamba achieves highly competitive performance against state-of-the-art methods. The source code is available at \url{https://github.com/JL021/U2Mamba}. 
\end{abstract}
\section{Introduction}
\label{sec:intro}

Salient Object Detection (SOD) is a fundamental problem in visual understanding, playing a critical role in applications such as medical imaging, autonomous driving, scene recognition, and video analysis~\cite{mo2022review}. The objective of SOD is to accurately identify visually prominent objects with precise boundaries. Although recent deep learning approaches have substantially advanced SOD performance, existing methods still face challenges in balancing global context modeling and fine-grained boundary preservation, particularly under high-resolution settings.

Most existing SOD methods are built upon convolutional neural networks (CNNs) or transformer-based architectures. CNN-based models, typically implemented as fully convolutional networks (FCNs)~\cite{villa2018fcn}, progressively downsample feature maps to reduce computational cost and enlarge receptive fields. However, this strategy often leads to the loss of high-resolution spatial details that are crucial for accurate saliency boundaries. Transformer-based models, such as Vision Transformer (ViT) and its variants~\cite{hassani2023neighborhood,liu2021swin}, are effective in capturing long-range dependencies through self-attention mechanisms. Nevertheless, their quadratic computational complexity with respect to input size limits their practicality for high-resolution and resource-constrained SOD applications, such as real-time video analysis or deployment on edge devices~\cite{yao2024ssnet}.

Recently, state space models (SSMs), particularly structured SSMs such as Mamba, have emerged as an efficient alternative for long-range dependency modeling~\cite{zhang2024survey}. By reformulating sequence modeling with hardware-aware designs, Mamba achieves linear or near-linear complexity with respect to sequence length, making it highly suitable for dense prediction tasks. Despite their promise, existing visual Mamba-based models are not specifically tailored for SOD. Architectures such as VMamba~\cite{liu2024vmamba} and U-Mamba~\cite{ma2024u} prioritize global dependency modeling but often adopt flattened token representations or generic U-shaped designs, which can weaken shallow high-resolution features essential for accurate saliency boundaries. Moreover, recent SOD-oriented Mamba models, e.g., MambaSOD~\cite{zhan2025mambasod}, mainly focus on multi-modal settings (e.g., RGB-D), while other U-shaped Vision Mamba variants~\cite{zheng2024u} are designed for different tasks and lack SOD-specific hierarchical supervision.

To address these limitations, we propose \textbf{U$^2$Mamba}, a novel nested two-level U-structured network specifically designed for single-modal salient object detection. Unlike prior works that adapt Mamba as a drop-in replacement for transformers or CNN backbones, U$^2$Mamba is constructed from scratch without relying on classification-pretrained backbones. At the core of our framework is a newly designed \emph{Multiscale Mamba U-Block} (MMUB), which embeds the Mamba mechanism into a compact U-shaped structure. By performing most operations on downsampled feature maps, MMUB efficiently captures intra-stage multiscale representations while preserving high-resolution features, significantly reducing computational redundancy and memory consumption.

At the top level, U$^2$Mamba adopts a U-Net-like encoder–decoder architecture, where each stage is composed of an MMUB. To further enhance representation learning and optimization stability, we introduce deep hierarchical supervision by applying loss functions at multiple levels. Specifically, we combine standard binary cross-entropy (BCE) loss with a Kullback–Leibler (KL) divergence-based probability matching loss, encouraging consistent saliency predictions across different semantic scales. In summary, the main contributions of this work are as follows:

1. We introduce \textbf{U$^2$Mamba}, a novel nested U-structured SOD framework that, for the first time, systematically integrates the Mamba state space mechanism into salient object detection.

2. We design a new \textbf{Multiscale Mamba U-Block (MMUB)}, enabling efficient intra-stage multiscale feature extraction with reduced spatial redundancy and memory cost, while preserving fine-grained saliency boundaries.

3. Extensive experiments demonstrate that U$^2$Mamba obtains favorable results across multiple benchmarks, validating the effectiveness and scalability of Mamba-based architectures for salient object detection.

\section{Method}
\label{sec:Method}

The goal of U$^2$Mamba is to introduce a nested U-structure model with an advanced state space model (SSM) for the Salient Object Detection task. The entire architecture of U$^2$Mamba is then presented, which includes a multiscale Mamba mechanism and a stacked U-Net-like structure. The detailed structure of U$^2$Mamba will be introduced in the following sections.

\subsection{Architecture Overview}

\begin{figure*}[t]
  \centering
  \includegraphics[width=0.8\textwidth]{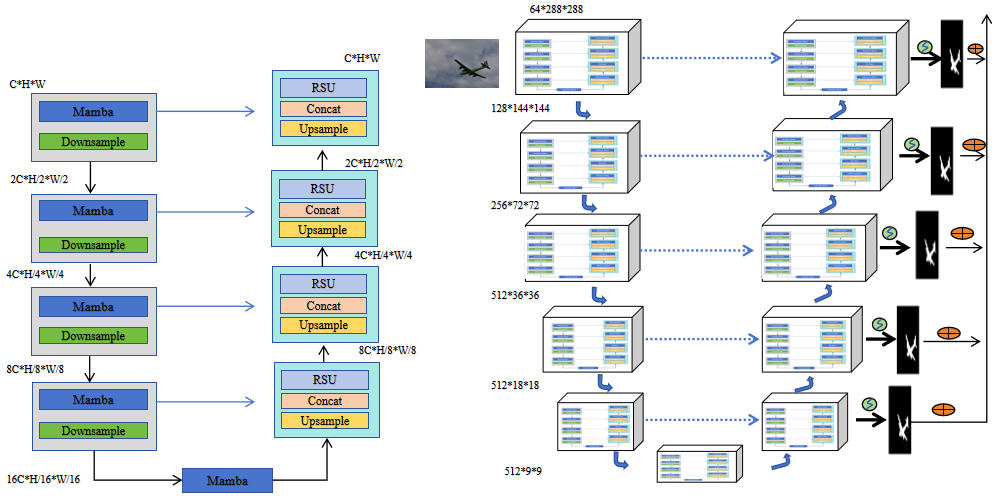}
  \caption{Illustration of the proposed model. (a) The MMUB enhances global context representation via deep multi-scale feature learning. (b) The overall architecture adopts a U-Net-like encoder–decoder structure, with each stage built upon our newly designed Multiscale Mamba U-Net Block (MMUB).
  }
  \label{fig:over}
  \vspace{-0.3cm}
\end{figure*}

U$^2$Mamba is a nested U-shaped encoder–decoder architecture that integrates Mamba state space modeling for efficient salient object detection. As shown in Fig.~\ref{fig:over}, the network follows a U$^2$Net-style design with six encoder stages ($En_1$–$En_6$) and five decoder stages ($De_1$–$De_5$), where saliency maps are predicted at multiple levels for deep supervision. Each stage is built upon a Multiscale Mamba U-Block (MMUB), which embeds the Mamba mechanism into a compact U-shaped structure to capture intra-stage multiscale features while preserving feature map resolution. By performing most computations on downsampled representations and explicitly retaining high-resolution shallow features, the nested design enables accurate boundary localization and efficient long-range context modeling. This architecture effectively addresses the dual requirements of SOD with significantly reduced computational and memory cost compared to attention-based models.

\subsection{Multiscales Mamba U Blocks}
Given that images are essentially discrete data sampled from continuous signals and can be treated as long sequences when flattened, we use Mamba’s linear scaling advantage to enhance CNNs’ long-range dependency modeling. In addition, both local and global contextual information are important for salient object detection and other segmentation tasks. Inspired by U-Mamba, we provide an innovative mamba-based U-structure block to utilize intrastage multiscale features to enhance detection efficacy. The MMUB, as illustrated in Fig.~\ref{fig:over}(a), adheres to the primary structure of RSU in U$^2$Net. 

The MMUB is designed to acquire the combination of best U-Net and Mamba for global context understanding in salient object detection. As shown in Figure~\ref{fig:over}, our MMUB is primarily composed of two components: a residual block (RSU) and a Mamba block. The residual block contains the plain convolutional layer, followed by Instance Normalization (IN). In the Mamba block, image features with a shape of $(B, C, H, W, D)$ are flattened and transposed to $(B, L, C)$, where $L = H \times W \times D$. This residual block transforms the input feature map $X(H \times W \times C_in)$ into an intermediate map $F_{1}(x)$ with a channel of $C_{out}$. A U-Net-like symmetric encoder-decoder structure with a height of $L$, which becomes deeper as $L$ increases. Finally, the features are projected back to the original shape $(B, L, C) $and then reshaped and transposed to $(B, C, H, W, D)$.

In each MMUB, $C_{\text{in}}$, $M$, and $C_{\text{out}}$ denote the input channel, intermediate channel, and output channel of the block, respectively. Higher encoder layers are designed to capture richer large-scale contextual information from feature maps, accompanied by gradual increases in feature height and width. Among the five encoder stages , $stage_3$ and $stage_4$ adopt dilated convolutions, which is motivated by the relatively low spatial resolution of feature maps at these two stages.

The decoder contains four stages and follows a symmetric architecture corresponding to the encoder. At each decoder stage, the input feature is formed by concatenating the upsampled feature map from the previous decoder stage with the skip feature map derived from its symmetric encoder counterpart.
The computational formulation of the Mamba Block is defined as:
\begin{equation}
F_c(f)=\sigma\big(\text{LN}(\text{ssm}(x))+x\big)
\end{equation}
where $x$ denotes the input feature of the Mamba Block, $\text{ssm}(\cdot)$ represents the structured state space modeling operation, $\text{LN}(\cdot)$ indicates layer normalization, and $\sigma(\cdot)$ denotes the nonlinear activation function.

\subsection{U$^2$Mamba}
Inspired by U$^2$-Net and U-Mamba, we propose a novel U$^2$-Mamba, where the exponential notation 2 indicates the level of the nested U structure. As illustrated in Fig.~\ref{fig:over}, each stage of U$^2$-Mamba is populated with a well-configured block consisting of 11 stages that collectively form a U structure. In general, U$^2$-Mamba consists of two main components.

Our approach preserves high-resolution characteristics while capturing multi-scale data at various stages. In several computer vision tasks, convolutional neural networks (CNNs) have demonstrated exceptional performance~\cite{wu2022uiu}. However, it is important to acknowledge the significant spatial redundancy that contributes to this high accuracy. A key innovation of MMUB is its strategy to mitigate the significant spatial redundancy inherent in CNN feature maps. It achieves this by decomposing the intermediate feature representation into high- and low-frequency components.  Low-frequency components with high spatial redundancy are processed at reduced resolution to save computation and memory, while high-frequency components containing critical edge and boundary information are fully preserved to maintain fine-grained saliency details. This frequency-based decomposition allows MMUB to reduce memory footprint and computational cost without compromising representational power, effectively enhancing both efficiency and performance. 
Furthermore, we successfully expanded the receptive field in pixel space by leveraging MMUB's capability to perform equivalent (low-frequency) convolution on low-frequency inputs. Compared to U$^2$-Net, our architecture maintains a deep structure with high resolution, while further improving segmentation accuracy and reducing computational and memory costs. Two MMUB blocks are utilized to address the bottleneck of U$^2$Mamba. Each level of the encoder and decoder employs skip connections to blend multiscale features with upscaled outputs, thereby enhancing spatial detail by merging shallow and deep layers. A subsequent linear layer preserves the dimensionality of this integrated feature set, ensuring consistency with the increased resolution. 


\subsection{Loss Function}
During training, we adopt a hierarchical deep supervision strategy inspired by U$^2$Net~\cite{2020U2}. While conventional deep supervision typically applies independent pixel-wise losses to side outputs, such schemes become less effective for very deep nested architectures due to weak inter-level consistency. To address this issue, we supervise each level of U$^2$Mamba using a combination of binary cross-entropy (BCE) loss and Kullback–Leibler (KL) divergence loss, following the spirit of~\cite{wang2020u2}.
The BCE loss $\mathcal{L}_{B}$ enforces pixel-level agreement with the ground-truth saliency map, whereas the KL divergence loss $\mathcal{L}_{K}$ explicitly aligns the probability distributions of predictions across different stages of the nested U-structure. By encouraging distribution-level consistency among hierarchical outputs, this supervision scheme facilitates coherent representation learning across depths, leading to more accurate boundary localization and robust saliency prediction.
The binary cross-entropy loss is:
\begin{equation}
    \mathcal L_{B}=-\sum^{H,W}_{i,j}[G_{(i,j)}logS_{(i,j)}+(1-G_{(i,j)}log(1-S_{(i,j)}]
\end{equation}

To promote multilevel interaction among different layers, we incorporate a pairwise probability prediction matching loss (KL loss) between any two levels. and the Kullback-Leibler divergence loss is:
\begin{equation}
    \mathcal L_{K}=-\sum^{H,W}_{i,j} G_{(i,j)}log \frac{G_{(i,j)}}{S_{(i,j)}}
\end{equation}
where $(i,j)$ are the pixel coordinates and $(H, W)$ is the height and width of the image. $G_(i,j)$ and $S_(i,j)$ denote the pixel values of the ground truth and the predicted moving object segmentation result, respectively. Our training loss is defined as:
\begin{equation}\label{eq:total_loss}
    Loss=\lambda_1 \sum_{i=1}^M \omega^{i}_{B} \mathcal L^i_{B}+\lambda_2 \sum_{i=1}^M  \omega^i_{K} \mathcal L^i _{K}+\lambda_3 \omega_{fuse}L_{fuse}
\end{equation}
where $\mathcal L_B$ and $\mathcal L_K$ (M = 6, as the side of each stage) denote the binary cross-entropy loss and the Kullback-Leibler divergence loss of the side output result, respectively. $\omega_{B}$ and $\omega_{K}$ are the weights of each loss term. $L_{fuse}$ is the loss of the final fusion output saliency, while $\omega_{fuse}$ represents its weights. We try to minimize the overall loss in the training process. 

\section{Experiments}
\label{sec:experiments}

\subsection{Datasets and Metrics}
We train our network on DUTS-TR, which has 10553 images in total. Currently, it is the largest and most popular training dataset for salient object detection. We supplemented this dataset with 21106 training photos obtained offline using horizontal flipping. We evaluate our method on several frequently used benchmark datasets, including ECSSD~\cite{yan2013hierarchical} with 1000 images, PASCAL-S~\cite{li2014secrets}  with 850 images, DUTOMRON~\cite{yang2013saliency} with 5168 images, HKU-IS~\cite{li2015visual} with 4,447 images, and DUTS~\cite{wang2017learning} with 15,572 images. All datasets are human-labeled with pixel-wise ground-truth for quantitative evaluations. We adopt three saliency metrics to evaluate the model performance, i.e., Mean Absolute Error (MAE), maximum F-measure ($maxF_{\beta}$)~\cite{ma2023boosting}.
\subsection{Implementation details}
During the object detection training phase, each image is first scaled to 320 by 320 pixels before being randomly flipped vertically and cropped to 288 by 288 pixels. Our network does not use any existing backbones. The weights for the $\omega^{i}$ and $\omega_{fuse}$ loss are set to one. For training, we utilized the Adam optimizer with the default hyperparameters (initial learning rate lr = 1e-3, betas = (0.9, 0.999), eps = 1e-4, and weight decay = 0. The hyperparameters for the composite loss function (Eq.~\ref{eq:total_loss}) are empirically set to $\lambda_1 = 0.4$, $\lambda_2 = 0.4$, and $\lambda_3 = 0.2$. The weights for each side output and the final fusion output are all set to one, i.e., $\omega_i^B = \omega_i^K = \omega_{\text{fuse}} = 1$. These values are used consistently across all experiments. Using earlier approaches, we train the network until the loss converges, without utilizing a validation set. The entire training process takes about 72 hours, and the training loss converges after 360 iterations (with a batch size of 16).  To obtain the saliency maps during testing, the image pictures are scaled to 320 by 320 pixels and then fed into the network. Both resizing methods utilize bilinear. Our network is built on PyTorch version 3.10.15, and we use A100 GPUs with 80GB for both training and testing.

\subsection{Comparison with State-of-the-Arts}
\begin{figure}[t]

	\centering
	\includegraphics[width=\linewidth]{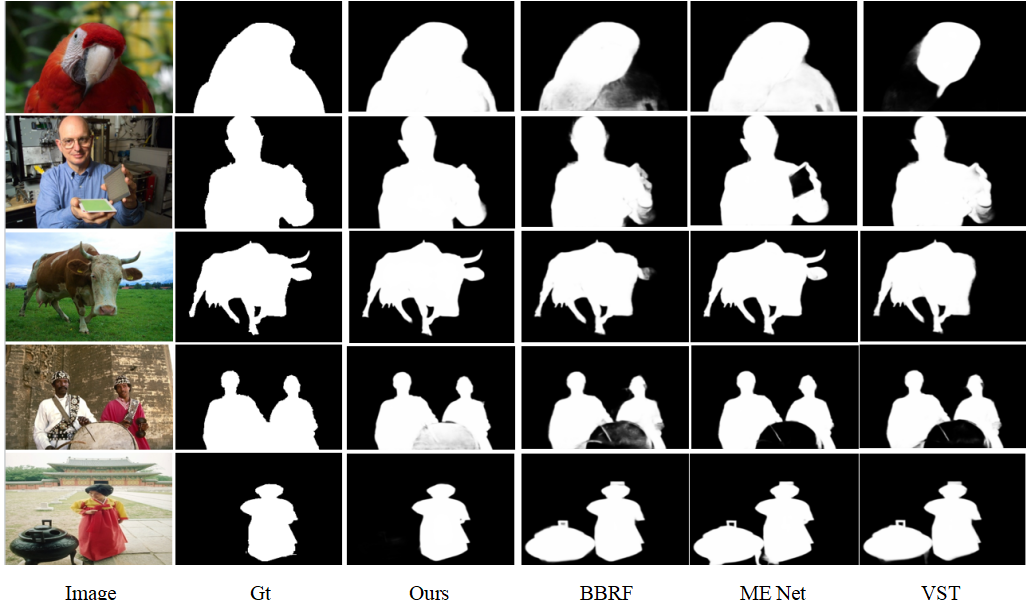}
	\caption{Visualization results on the ECSSD dataset. }
    \label{sam}
	 \vspace{-6mm}
\end{figure}

For comparative purposes, all baseline methods were trained using identical hyperparameter configurations. Ten cutting-edge saliency detection models: F3Net~\cite{wei2020f3net}, RCSB~\cite{ke2022recursive}, U$^2$Net~\cite{2020U2}, MSENET~\cite{lee2022msenet}, LDF~\cite{wei2020label}, POOLNET~\cite{rani2023poolnet}, BBRF~\cite{ma2023boosting}, MENET~\cite{wang2023pixels}, VST~\cite{liu2021visual}, and VST-S++~\cite{liu2024vst++} are compared to the proposed framework. The comparative results are presented in Tables~\ref{result}. These tables demonstrate that our method's MSE and $maxF_{\beta}$ surpass or close those of the currently employed techniques. Furthermore, our approach achieves the lower MAE, indicating that our predictions are the most accurate. The method proposed in this paper is well correlated with the detection of salient objects. A comparison of the results in Figure~\ref{sam} reveals that the proposed algorithm consistently generates more accurate and comprehensive saliency maps, boundaries, and distinct borders, and also aligns more closely with the ground truth map methods. U$^2$Mamba generates sharper boundaries and more coherent salient regions, especially in complex scenes with cluttered backgrounds or multiple objects. However, like most SOD methods, it can struggle when the salient object shares strong textural or color similarity with the background. Future work could integrate explicit edge guidance or semantic priors to address such ambiguities. The approach suggested in this research effectively reduces background, emphasizing and highlighting the significant area in images. More results can be found in ~\href{https://github.com/JL021/U2Mamba}{https://github.com/JL021/U2Mamba}.

\begin{table}[h]
\caption{Comparison of our method and SOTA methods on baseline datasets }
\vspace{-0.4cm}
\label{result}
     \begin{center}
\begin{small}
\begin{sc}
\setlength{\tabcolsep}{+0.1mm}{
\scalebox{0.8}{\begin{tabular}{lcccccccccc}
\toprule
\multirow{2}{*}{Model}  &\multicolumn{2}{c}{DUTS-TE}&\multicolumn{2}{c}{PASCAL} &\multicolumn{2}{c}{DUT-OMRON}&\multicolumn{2}{c}{HKU-IS}&\multicolumn{2}{c}{ECSSD}\\
\cline{2-11}
~& $maxF_{\beta}$ & MAE&$maxF_{\beta}$ & MAE&$maxF_{\beta}$ & MAE&$maxF_{\beta}$ & MAE&$maxF_{\beta}$ & MAE  \\
\midrule
F3Net&0.840&0.035&0.840&0.062&0.766&0.053&0.910&0.028&0.925&0.033\\
RCSB&0.855&0.034&0.842&0.058&0.773&0.045&0.923&0.027&0.823&0.033\\
U$^2$Net&0.873&0.044&0.770&0.076&0.823&0.054&0.935&0.031&0.951&0.033\\
MSENet&0.877&0.034&0.862&0.060&0.798&0.045&0.927&0.026&0.941&0.033\\
LDF&0.855&0.034&0.848&0.060&0.773&0.051&0.914&0.027&0.930&0.034\\
PoolNet&0.809&0.040&0.822&0.074&0.747&0.056&0.899&0.032&0.915&0.039\\
BBRF&0.905&0.040&0.884&0.074&0.820&0.056&0.946&0.032&0.957&0.039\\
MENet&0.895&0.028&0.848&0.062&0.792&0.045&0.939&0.023&0.938&0.031\\
VST&0.877&0.037&0.850&0.067&0.800&0.058&0.937&0.030&0.944&0.034\\
VST-S++&0.897&0.029&0.859&0.062&0.813&0.050&0.941&0.025&0.951&0.027\\
\hline
U$^2$Mamba     &0.904 &0.024&0.856&0.068& 0.816& 0.052& 0.933 &0.025&0.929&0.024 \\
\bottomrule
\end{tabular}}}
\end{sc}
\end{small}
\end{center}

\end{table}

\vspace{-1mm}
\subsection{Ablation Study}
To evaluate the effectiveness of each core component in U$^2$Mamba, we perform ablation experiments on the DUTS-TE dataset with consistent training configurations. As reported in Table~\ref{tab:ablation}, we first construct a solid baseline (U$^2$Net-RSU) by utilizing original Residual U-blocks (RSUs) in the nested U-structure, which yields a maxF$_\beta$ of 0.873 and MAE of 0.044. First, enhancing the baseline with our hierarchical supervision (BCE + KL divergence) brings obvious performance gains (maxF$_\beta$: 0.883, MAE: 0.035), demonstrating that KL divergence promotes feature consistency and boundary precision across levels. Furthermore, replacing RSUs with our proposed MMUB also achieves significant performance improvements (maxF$_\beta$: 0.891, MAE: 0.031), verifying the superiority of multi-scale Mamba feature learning for salient object detection. Finally, the full model equipped with both MMUB and hierarchical supervision achieves the optimal results (maxF$_\beta$: 0.904, MAE: 0.024). These results validate that the integration of MMUB and hierarchical supervision is essential for the high performance of our method, with each component independently contributing to the model's efficacy.

\begin{table}[!t]
\centering
\vspace{-3mm}
\caption{Ablation study on DUTS-TE.}
\label{tab:ablation}
\begin{tabular}{lcc}
\toprule
Model Variant & maxF$_\beta$ $\uparrow$ & MAE $\downarrow$ \\
\midrule
U$^2$Net-RSU (Baseline) & 0.873 & 0.044 \\
+BCE+KL & 0.883 & 0.035 \\
+MMUB& 0.891 & 0.031 \\
\textbf{Full U$^2$Mamba} & \textbf{0.904} & \textbf{0.024} \\
\bottomrule
\end{tabular}
\vspace{-2mm}
\end{table}

\vspace{-1mm}
\subsection{Computational Efficiency Analysis}
We compare the computational overhead of U$^2$Mamba  against representative CNN-based (U$^2$Net and Transformer-based (VST) ) SOD models. As shown in Table~\ref{tab:efficiency}, U$^2$Mamba achieves a favorable trade-off between accuracy and efficiency. Despite having fewer parameters (41.97M vs. $U^2Net$'s 176.3M) and lower FLOPs (127.5G vs. VST's 198.2G), it outperforms both in $maxF_\beta$. Moreover, thanks to the linear complexity of the Mamba block with respect to sequence length, U$^2$Mamba exhibits faster inference speed (24.7 FPS on an A100 GPU) compared to VST (18.3 FPS), making it more suitable for real-time applications.

\begin{table}[!t]
\centering
\vspace{-2mm}
\caption{Computational efficiency comparison on DUTS-TE.}
\label{tab:efficiency}
\begin{tabular}{lccc}
\toprule
Model & Params (M) & FLOPs (G) & FPS \\
\midrule
$U^2Net$& 176.3 & 142.8 & 26.1 \\
VST& 94.7 & 198.2 & 18.3 \\
\textbf{$U^2Mamba $(Ours)} & \textbf{41.97} & \textbf{127.5} & \textbf{24.7} \\
\bottomrule
\end{tabular}

\end{table}

\vspace{-1mm}
\subsection{Effectiveness of Mamba Mechanism}
To further justify the choice of Mamba over other long-range modeling strategies, we replace the core SSM module in MMUB with either a standard self-attention layer (Transformer-style) or a dilated convolution stack (CNN-style). The results in Table~\ref{tab:mamba_vs} show that the Mamba variant consistently outperforms its counterparts, achieving the highest maxF$_\beta$ and lowest MAE. This highlights Mamba's unique advantage in efficiently modeling global context with linear complexity, which is particularly beneficial for high-resolution saliency maps where quadratic attention becomes prohibitive.

\begin{table}[!t]
\centering
\vspace{-5mm}
\caption{Comparison of different long-range modeling mechanisms within MMUB on DUTS-TE.}
\label{tab:mamba_vs}
\begin{tabular}{lcc}
\toprule
Long-range Module & maxF$_\beta$ $\uparrow$ & MAE $\downarrow$ \\
\midrule
Dilated Convolution & 0.887 & 0.032 \\
Self-Attention & 0.895 & 0.029 \\
\textbf{Mamba (SSM)} & \textbf{0.904} & \textbf{0.024} \\
\bottomrule
\end{tabular}
\end{table}

\vspace{-3mm}
\section{Conclusion}
\label{sec:conclusion}

In this research, we present a multi-scale Mamba mechanism for salient object detection utilizing a novel nested U structure, referred to as U$^2$-Mamba. Our developed multiscale Mamba U structure block (MMUB) is integrated into each stage of U$^2$-Mamba. This design enables U$^2$-Mamba to capture rich local and global information at high resolution while reducing computational resources and spatial redundancy compared to the original U$^2$Net's CNN. We propose a hierarchical training supervision approach to enhance the efficacy of the deep network. Experimental results on several mainstream salient object detection datasets demonstrate that the proposed method achieves competitive performance against existing approaches.


\small
\bibliographystyle{IEEEbib}
\bibliography{strings,refs}

\end{document}